\documentclass{article}

\PassOptionsToPackage{numbers,sort}{natbib}

\usepackage[preprint]{neurips_2024}

\usepackage[numbers]{natbib}
\usepackage[utf8]{inputenc} 
\usepackage[T1]{fontenc}    
\usepackage{hyperref}       
\usepackage{url}            
\usepackage{booktabs}       
\usepackage{amsfonts}       
\usepackage{nicefrac}       
\usepackage{microtype}      
\usepackage{xcolor}         
\usepackage{wrapfig}
\usepackage{adjustbox}
\usepackage{multirow}

\title{Can Language Models Evaluate Human Written Text? Case Study on Korean Student Writing for Education}

\author{
Seungyoon Kim\\
  Sangmoon High School\\
  \texttt{seungyoon2007@gmail.com} \\
  \And
  Seungone Kim \\
  Carnegie Mellon University \\
  \texttt{seungone@cmu.edu} \\
}

\begin{document}

\maketitle

\begin{abstract}
Large language model (LLM)-based evaluation pipelines have demonstrated their capability to robustly evaluate machine-generated text. Extending this methodology to assess human-written text could significantly benefit educational settings by providing direct feedback to enhance writing skills, although this application is not straightforward. In this paper, we investigate whether LLMs can effectively assess human-written text for educational purposes. We collected 100 texts from 32 Korean students across 15 types of writing and employed GPT-4-Turbo to evaluate them using grammaticality, fluency, coherence, consistency, and relevance as criteria. Our analyses indicate that LLM evaluators can reliably assess grammaticality and fluency, as well as more objective types of writing, though they struggle with other criteria and types of writing. We publicly release our dataset and feedback~\footnote{\href{https://github.com/seungyoon1/llm-as-a-judge-human-eval/tree/main}{https://github.com/seungyoon1/llm-as-a-judge-human-eval/tree/main}}.
\end{abstract}

\section{Introduction}
Recent works have demonstrated that when adeptly prompted or trained, large language models (LLMs) can closely mimic human evaluation in assessing machine-generated text~\citep{zheng2023judging,alpaca_eval,liu2023gpteval}. This has led to the development of numerous generative evaluation pipelines that use language models as judges (denoted as `LLM-as-a-Judge')~\citep{kim2023prometheus,verga2024replacing,dubois2024length}. Applying this methodology to human-written text could greatly benefit educational settings by providing direct feedback to enhance writing skills~\citep{henderson2019challenges,wisniewski2020power}. However, human-written text often possesses different characteristics from machine-generated text, particularly in the unexpected use of words~\citep{holtzman2019curious}. This discrepancy complicates the direct application of the LLM-as-a-Judge pipeline for evaluating human-written text. In this work, we explore whether the LLM-as-a-Judge pipeline can also be used to evaluate \textit{human-written text}. Specifically, we are interested in determining if LLMs can effectively assess texts written by Korean students to improve their writing skills across various formats, such as essays, reports, and scripts.

For this study, we collected a corpus of 100 texts written by 32 Korean students, ranging in age from 11 to 19, covering 15 types of writing. We evaluated these texts using GPT-4-Turbo based on five criteria: `grammaticality,' `fluency,' `coherence,' `consistency,' and `relevance,' resulting in a total of 500 judgments across the 100 texts. Subsequently, we redistributed the verbal feedback generated by GPT-4-Turbo to the students, asking them to assess whether the judgments were reasonable, overly critical, or overly optimistic. The results indicated that while judgments on `grammaticality' and `fluency' were mostly deemed reasonable (87\% and 93\%, respectively), those concerning the other three criteria were relatively lower, suggesting that LLM-as-a-Judge pipeline has limitations. Notably, upon examining individual types of writing, participants found the judgments to be particularly unreasonable when assessing more subjective texts, such as diaries and self-introduction essays.

Although it was not a rigorous comparison of writings on the same topic in a controlled environment, our analysis of the evaluation results allowed us to make several indirect observations that LLM-as-a-Judge could be utilized by students to produce higher quality writing. First, we observed that while it tends to give high scores for Consistency and Relevance, it gives lower scores for Fluency. From this, we were able to indirectly observe through the participants' assessments that found the results reasonable, that LLM-as-a-Judge could be used in educational settings to help students write more fluently. Secondly, students often wrote descriptive essays and book reports, and the evaluation results for these writing types were found to be reasonably high. Most scores across the five criteria ranged from the high 3s to the low 4s, suggesting that LLM-as-a-Judge could be useful in elevating writings from a 3.5 - 4 out of 5 to a full 5. Lastly, when comparing the average scores between texts written by 11-13-year-olds and those by 17-19-year-olds, we found that the latter tended to receive higher scores. This ability to discriminate between the two age groups not only confirmed the robustness of LLM-as-a-Judge but also suggested that it could be actively used to help younger students elevate the quality of their writing to that of older students.


\section{Related Works}

Due to the ambiguity in grading free-form responses generated by LLMs, word-match or semantic-similarity based evaluation metrics have long been the de-facto standards~\citep{lin2004rouge,papineni2002bleu,zhang2019bertscore}. Recently, directly employing language models as evaluators (LLM-as-a-Judge) has proven more effective, achieving a higher correlation with human grading of machine-generated text~\citep{zheng2023judging,alpaca_eval,liu2023gpteval,chan2023chateval,kim2023prometheus,dubois2024length,verga2024replacing}. Moreover, follow-up studies have demonstrated that employing detailed evaluation criteria not only yields more accurate judgments but also enhances interpretability~\citep{ye2023flask,kim2023evallm,kim2023prometheus,kim2024prometheus,lee2024prometheus,kim2024biggen}. Inspired by this, our work employs five specific evaluation axes: grammaticality, fluency, coherence, consistency, and relevance. On the application side, prior works have used LLM-as-a-Judge to evaluate various types of LLM-generated text, including the soundness of research proposals~\citep{shao2024assisting}, toxicity in life advice~\citep{kim2023lifetox}, and the factuality of medical documents~\citep{joseph2024factpico}. Our work distinguishes itself from these examples as we evaluate \textit{human-written text} across 15 distinct types of writing from students.

In the field of adopting LLMs to enhance education, prior works have primarily focused on distinguishing between LLM-generated and human-generated text~\citep{orenstrakh2023detecting,bernabei2023students,gao2024llm}. While preventing the overuse of LLMs is crucial, it is equally important to explore better educational models for using LLMs wisely. Our work suggests that LLMs can be effectively utilized as tools to enhance students' writing by providing verbal feedback on their writing.

\section{Experimental Setting}

In this section, we describe (1) the process by which we gathered and preprocessed 100 human-written texts, (2) our assessment pipeline for acquiring judgments from LLM evaluators, and (3) how we asked each student to determine whether the judgments from LLM evaluators were reasonable.

\subsection{Dataset Construction}\label{sec:3.1}

\begin{wraptable}{R}{0.36\textwidth}
\vspace{-6mm}
\centering
\caption{Statistics of age distribution of the 100 student-written texts.}
\vspace{3mm}
\label{tab:age_dist}
\fontsize{8}{10}\selectfont
\begin{tabular}{cc}
    \toprule
    \textbf{Age}            & \textbf{Number of Instances}\\ \midrule
    11 & 16\\
    13 & 18\\
    14 & 4\\
    17 & 33\\
    18 & 26\\
    19 & 3\\
    \bottomrule
\end{tabular}
\vspace{-3mm}
\end{wraptable}

We gathered 32 participants, all of whom are Korean students. Each participant was asked to provide a free-form text that they had manually written without using LLMs, along with instructions for writing that text. This includes writings from the past, for which we labeled the age at which the text was written. We excluded texts that were part of classification tasks or were short answers. After this process, the main authors categorized the text into 15 types of writing, including Book Report, Descriptive Essay, Process Essay, Reflective Essay, Story Writing, Play Script, Linguistic Report, Scientific Report, Presentation Script, Problem Creation, Argumentative Essay, Presentation Report, Diary, Self Introduction Essay, and Letter.

The main authors manually typed and revised the texts based on the instructions, as these documents are typically long and in PDF format. To maintain the quality of the text, we did not revise the content or grammar. Instead, we manually corrected line break issues using an online JSON formatting tool~\footnote{\href{https://jsoneditoronline.org/}{https://jsoneditoronline.org/}} to ensure compatibility of the dataset. Each instance consists of six components: student id (ranging between 0 and 31 to distinguish the writer of each text), age (ranging between 11 and 19, based on the age when the writer composed the text, not their current age), language (either Korean or English), type of writing (among the 15 categories listed above), input (the instruction to write the corresponding text), and output (the text that each student wrote). The statistics of the age, type of writing, and language are shown in Table~\ref{tab:age_dist}, Table~\ref{tab:language_dist}, and Table~\ref{tab:type_dist}.

\begin{wraptable}{R}{0.47\textwidth}
\vspace{-6mm}
\centering
\caption{Statistics of age distribution of the 100 student-written texts.}
\vspace{3mm}
\label{tab:type_dist}
\fontsize{8}{10}\selectfont
\begin{tabular}{lc}
    \toprule
    \textbf{Type of Writing}            & \textbf{Number of Instances}\\ \midrule
    Book Report & 17\\
    Descriptive Essay & 22\\
    Process Essay & 4\\
    Reflective Essay & 7\\
    Story Writing & 1\\
    Play Script & 3 \\
    Linguistic Report & 1\\
    Scientific Report & 13\\
    Presentation Script & 9\\
    Problem Creation & 2 \\
    Argumentative Essay & 11 \\
    Presentation Report & 2 \\
    Diary & 5\\
    Self Introduction Essay & 2\\
    Letter & 1\\
    \bottomrule
\end{tabular}
\vspace{-3mm}
\end{wraptable}

\subsection{Evaluation Pipeline}\label{sec:3.2}

We mainly follow the evaluation pipeline of Prometheus~\citep{kim2023prometheus,kim2024prometheus}. Specifically, we utilize the \texttt{prometheus-eval} library~\footnote{\href{https://github.com/prometheus-eval/prometheus-eval}{https://github.com/prometheus-eval/prometheus-eval}}, following the default hyper-parameters, and add five custom score rubrics to evaluate texts written by students. The code and prompts are provided in supplementary material. We use GPT-4-Turbo-April as the evaluator for conducting 500 judgments. The judgments consist of verbal feedback, which highlights the strengths and weaknesses of the student's text with respect to each evaluation criterion, and a scoring decision, which is an integer between 1 and 5. The criteria are as follows: the fluency criterion assesses if the text fluent and easy to read, considering it is written by a Korean student. The coherence criterion evaluates whether the text is coherent and logically organized, considering it is written by a Korean student. The consistency criterion examines if the text is consistent in terms of style, tone, and tense. The relevance criterion checks if the text is relevant to the given instruction or topic. Lastly, the grammaticality criterion assesses whether the text demonstrates proper grammatical usage.

\begin{wraptable}{R}{0.42\textwidth}
\vspace{-6mm}
\centering
\caption{Statistics of age distribution of the 100 student-written texts.}
\vspace{3mm}
\label{tab:language_dist}
\fontsize{8}{10}\selectfont
\begin{tabular}{cc}
    \toprule
    \textbf{Written Language}            & \textbf{Number of Instances}\\ \midrule
    Korean & 63\\
    English & 37\\
    \bottomrule
\end{tabular}
\vspace{-3mm}
\end{wraptable}

\subsection{Verification of Evaluation Results}\label{sec:3.3}

To verify if the evaluation results from Subsection~\ref{sec:3.2} are valid, we distribute the verbal feedback and scoring decisions back to the 32 student participants who provided their writing. We ask if the scoring decisions and the verbal feedback were valid or not (\textit{i.e.}, the scoring as well as the verbal feedback is overly critical or overly optimistic). Each participant were paid \$ 10 for providing each writing piece and verifying the judgments (15 minutes per instance). Although we were unable to confirm whether students' writing improved through direct revisions due to cost limitations, the goal was to indirectly verify if they could improve their writing by reviewing whether the feedback was valid. Future work could involve experimenting with students directly revising their writing.

\section{Experimental Results \& Discussions}

\begin{table*}
\small
\caption{\textbf{GPT-4 evaluation results of 100 student-written texts across 5 evaluation criteria, and posthoc human meta-evaluation results on GPT-4's judgment.} Across all criteria, humans verified that on 77\% - 93\% of the time, the judgments were reasonable, which supports the claim that LLM-as-a-Judge could be utilized to pinpoint the strength and weakness of each student's writing.}
\vspace{1mm}
\centering
\resizebox{1.0\textwidth}{!}{
    \begin{tabular}{@{}lccccccccc@{}}
    \toprule
    \multicolumn{1}{c}{\multirow{2}{*}{\textbf{Criteria}}}& \multicolumn{6}{c}{\textsc{LLM-as-a-Judge Evaluation Results}} & \multicolumn{3}{c}{\textsc{Human Meta-Evaluation Results}}\\ 
    \cmidrule(lr){2-7} \cmidrule(lr){8-10} &
    \textbf{Score 1} & \textbf{Score 2}& \textbf{Score 3}& \textbf{Score 4}& \textbf{Score 5}& \textbf{Avg Score}& \textbf{Valid}& \textbf{Overly Critical}& \textbf{Overly Optimistic}\\ \midrule
    Grammaticality & 0 & 6 & 28 & 42 & 11 & 3.63 & 87\% & 8\% & 5\%\\
    Fluency & 0 & 3 & 72 & 23 & 2 & 3.24 & 93\% & 6\% & 1\%\\
    Coherence & 1 & 4 & 34 & 39 & 22 & 3.77 & 81\% & 13\% & 6\%\\
    Consistency & 1 & 1 & 20 & 53 & 25 & 4.00 & 82\% & 4\% & 14\%\\
    Relevance & 7 & 9 & 13 & 30 & 41 & 3.89 & 77\% & 4\% & 19\%\\
    \bottomrule
    \end{tabular}
}\label{table:main}
\vspace{-3mm}
\end{table*}

\paragraph{Can LLM-as-a-Judge provide reasonable judgments?} The GPT-4 evaluation results and post-hoc human meta-evaluation results are shown in Table~\ref{table:main} (on the next page). First, when examining GPT-4's scoring decisions for the same fixed 100 texts, consistency and relevance received higher scores, while fluency received lower scores. Second, in the meta-evaluation results conducted by the students, fluency had the highest agreement on the validity of the judgment. Given that fluency achieved the lowest score and humans determined it to be the most valid (93\%), this supports the claim that the LLM-as-a-Judge pipeline could be utilized to enhance students' writing. This finding is straightforward, considering that the participants we tested are all Korean students who do not use English as their first language. Also, for more objective evaluation criteria such as grammaticality, the validity was high (87\%).

\begin{wraptable}{R}{0.40\textwidth}
\centering
\caption{\textbf{Human meta-evaluation results on GPT-4's judgment for each writing category.} Validity ratio is higher on more objective texts while lower on subjective texts.}
\vspace{3mm}
\label{tab:validity_ratio}
\fontsize{8}{10}\selectfont
\begin{tabular}{lc}
    \toprule
    \textbf{Type of Writing}            & \textbf{Ratio of Validity}\\ \midrule
    Process Essay & 100.00\%\\
    Descriptive Essay & 95.45\%\\
    Scientific Report & 92.31\%\\
    Self Introduction Essay & 50.00\%\\
    Argumentative Essay & 36.36\% \\
    Diary & 20.00\%\\
    \bottomrule
\end{tabular}
\vspace{-3mm}
\end{wraptable}

\paragraph{Is there a type of writing where LLM-as-a-Judge does not function properly?} Considering that our text collection consists of a wide variety of writings, we further analyzed the patterns of which writings were deemed to consist of more accurate judgments. Table~\ref{tab:validity_ratio} shows the top 1, 2, and 3 and lowest 1, 2, and 3 validity ratios for the writing categories (excluding categories with only one text). The validity ratio tends to be higher for texts with more objective characteristics, including process essays, descriptive essays, and scientific reports, while it is lower for self-introduction essays, argumentative essays, and diaries. Notably, we observed that for subjective tasks, texts are often written in a more informal tone. However, coherence and consistency criteria, which should not consider informality, pointed this out as a limitation, indicating that LLM-as-a-Judge does not function as intended in these cases. We leave the development of better evaluation frameworks to assess human-written text to assist their writing for future work.

\begin{wraptable}{R}{0.40\textwidth}
\vspace{-7mm}
\centering
\caption{\textbf{GPT-4 evaluation results for each age group across 5 criteria.} Higher scores are given to senior students over junior students.}
\vspace{3mm}
\label{tab:age_group_dist}
\fontsize{8}{10}\selectfont
\begin{tabular}{ccc}
    \toprule
    \textbf{Criteria}            & \textbf{Age 11-14} & \textbf{Age 17-19}\\ \midrule
    Grammaticality & 3.34 & 3.95\\
    Fluency & 3.08 & 3.25\\
    Coherence & 3.24 & 3.95\\
    Consistency & 3.53 & 4.23\\
    Relevance & 3.54 & 3.95\\
    \bottomrule
\end{tabular}
\vspace{-3mm}
\end{wraptable}

\paragraph{Is GPT-4 capable of distinguishing between writings from senior and junior students?} The average scores provided by GPT-4-Turbo evaluator is shown in Table~\ref{tab:age_group_dist}. Although it is not strictly accurate since the participants' writings were not compared based on the same instructions, we generally observed a difference in scores between the two groups across all criteria. This indicates that the LLM-as-a-Judge pipeline tends to give higher scores to senior students. This finding indirectly supports the reliability of LLM-as-a-Judge, based on the premise that older students are capable of writing more flexibly.

\section{Conclusion}

Recently, LLM-as-a-Judge has emerged as an effective paradigm for evaluating LLM responses to subjective questions. In this work, we extend this paradigm to investigate whether it can also effectively evaluate human-written texts through preliminary experiments. We collected and evaluated 100 essays written by a total of 32 Korean students of various ages, and then had each student verify whether the evaluation results were reasonable. The experimental results confirmed that LLM-as-a-Judge could serve as a good evaluator for more objective types of writing, as well as for fluency and grammaticality criteria. Additionally, we identified areas that need further improvement. We hope that our work will serve as a foundation for future research exploring the use of LLM-as-a-Judge to evaluate human-written texts more broadly.

\section*{Acknowledgement}

The first author, currently a highschool student, have proposed the idea for this paper, conducted all the experiments, and wrote the paper. The corresponding author have provided feedback throughout the project by having a weekly 2 hour meeting. 

\bibliography{neurips_2024}

\end{document}